\documentclass[times, 10pt]{article}
\usepackage{graphicx}
\usepackage{caption}
\usepackage[affil-it]{authblk}

\newtheorem{theorem}{Theorem}[section]\def\TH{
\begin{theo}}\def\HT{
\end{theo}}
\newtheorem{defi}[theorem]{Definition}\def\DE{
\begin{defi}}\def\ED{
\end{defi}}

\newtheorem{definition}{Definition}
\newtheorem{lemma}{Lemma}

\everymath{\displaystyle}
 
\begin{document}

\title{A new approach in machine learning\\ (Preliminary report)}
\author{
   Alain Tapp 
   \thanks{Electronic address: \texttt{alain.tapp@gmail.com}}
	}
\affil{Universit\'e de Montr\'eal}

\maketitle

\begin{abstract}

In this technical report we presented a novel approach to machine learning. 
Once the new framework is presented, we will provide a simple and yet very powerful learning algorithm which will be benchmark on various dataset. 

The framework we proposed is based on booleen circuits; more specifically the classifier produced by our algorithm have that form. 
Using bits and boolean gates instead of real numbers and multiplication enable the the learning algorithm and classifier to use very efficient boolean vector operations. 
This enable both the learning algorithm and classifier to be extremely efficient. 
The accuracy of the classifier we obtain with our framework compares very favorably those produced by conventional techniques, both in terms of efficiency and accuracy. 
\end{abstract}

%%%%%%%%%%%%%%%%%%%%%%%%%%%%%%%%%%%%%%%%%%%%%%%%%%%%%%%%%%%%%%%%%%%%%%%%%%%%%%%%%%%%%%%%%%%%%%%%%%%%%%%%%%%%%%%%%%%%%%
%%%%%%%%%%%%%%%%%%%%%%%%%%%%%%%%%%%%%%%%%%%%%%%%%%%%%%%%%%%%%%%%%%%%%%%%%%%%%%%%%%%%%%%%%%%%%%%%%%%%%%%%%%%%%%%%%%%%%%
\section*{Introduction}

This technical report is a draft and is not intended for publication in an official venue as is. 
Our goal it to share preliminary results in order to obtain feedback and guidance.  
Before making a big move to machine learning, the author has worked on cryptography, quantum computing, and complexity theory, amongst other things. 
In all of these fields, boolean circuits are at the heart of many of the deepest results. 
For one coming from these fields, it is surprising that they do not hold a prominant place in machine learning. 

In this note, we present a machine learning approach based on boolean circuits and study the characteristics of a learning algorithm that naturally fit this approach.
Our original motivation was unsupervised learning and although we firmly believe this approach can lead to interesting results in that field, 
we have decided to concentrate first on supervised classification.

Once one is acquainted with the framework, a straightforward supervised learning algorithm for binary classification emerges naturally. 
Studying the properties of this simple algorithm we find its performance impressive, both in terms of efficiency and accuracy. 
It compares very favorably with mainstream techniques like neural nets, deep neural nets and support vector machines. 
This is despite the relative maturity of these established techniques, compared with our new framework.   

The framework will be presented in the next section. We then present a straightforward greedy algorithm for classification. 
The following section adds hill climbing to the greedy algorithm, for a general learning algorithm that is our focus in this article. 
The conclusion contains a long list of area of improvement to the techniques presented in this note. 
Finally, the appendix contains detailed analysis of the behavior and performance of the algorithm on several benchmarks problems. 
These benchmark problems vary from numerical data to color images; of course MNIST is also part of the benchmarks.
Some of those benchmarks have been used in \cite{Larochelle:2007} and we can thus compare ourselves to the state of the art in 2007.

Once the technique is refined and especially once deep learning has been integrated into the model, we are optimistic this approach can compete with the best techniques available today.
We believe the fact that we obtain such impressive results, with very varied benchmarks, and with no engineering, is outstanding; and we believe this preliminary note provides justification and motivation for the further study of this paradigm.

%%%%%%%%%%%%%%%%%%%%%%%%%%%%%%%%%%%%%%%%%%%%%%%%%%%%%%%%%%%%%%%%%%%%%%%%%%%%%%%%%%%%%%%%%%%%%%%%%%%%%%%%%%%%%%%%%%%%%%
%%%%%%%%%%%%%%%%%%%%%%%%%%%%%%%%%%%%%%%%%%%%%%%%%%%%%%%%%%%%%%%%%%%%%%%%%%%%%%%%%%%%%%%%%%%%%%%%%%%%%%%%%%%%%%%%%%%%%%
\section*{The framework}

In this section we describe and formalize the new framework we propose. 
In contrast with approaches that use reals numbers, multiplication, and exponentiation, we present a framework based on binary number (bits) and boolean circuits.
In short, inputs are are binary vectors of a given length, and classifiers are boolean circuits.

The input data could be images, text, lists of numbers or anything else encoded as fixed-length binary vectors.  The classifier we will obtain by our algorithm will be booleen circuits, specifically tree-structured circuits. 
All the classifiers we present in this note are binary circuits with binary vector inputs, and output a single bit representing the classification decision. 

\begin{definition}
In this note, a $k$-gate is a function mapping $k$ inputs bits to an output bit. 
We call $k$ the arity of the gate.
\end{definition}

\begin{lemma}
The number of possible $k$-gate is $2^{2^k}$ and it can be completely specified by a look up table of $2^k$ bits.
\end{lemma}

For example, the boolean AND, OR and XOR gate are $2$-gate. 
There is a total of 16 boolean binary gates with 2 inputs (if one consider the input bit an order pair).  
The total number of possible gates for a given arity is surprisingly high.  
In the case of arity 8, 
there are 
\[ 115792089237316195423570985008687907853269984665640564039457584007913129639936 \] 
different gates that can be computed, 
but each of those is uniquely define by a truth table of 256 bits. 

\begin{figure}[ht]
\centering
\begin{minipage}[b]{0.45\linewidth}

\begin{center}
  \begin{tabular}{|| c|c|| }
    \hline\hline
     00&  0\\ \hline
    01 & 0\\ \hline
    10 & 0\\ \hline
    11 & 1\\ \hline
    \hline
  \end{tabular}
   \captionof{table}{2-gate, AND}
\end{center}

\label{fig:minipage1}
\end{minipage}
\quad
\begin{minipage}[b]{0.45\linewidth}

\begin{center}
  \begin{tabular}{|| c|c|| }
    \hline\hline
       000&  0\\ \hline
       001 & 0\\ \hline
       010 & 0\\ \hline
       011 & 1\\ \hline
      100&  0\\ \hline
      101 & 1\\ \hline
      110 & 1\\ \hline
      111 & 1\\ \hline
    \hline
  \end{tabular}
   \captionof{table}{3-gate, Majority}
\end{center}

\label{fig:minipage2}
\end{minipage}
\end{figure}

\begin{definition} 
In this note we define a {\em Boolean circuit classifier} as a Boolean circuit whose input bits are data bit and who's output bit represent the classification decision. 
\end{definition}

We often restrict the gates in a classifier to have the same arity, as a design choice.  All the circuits we consider in this note will be full trees (i.e., every leaf has the same distance from the root (in edges)).

\begin{definition}
The depth of a circuit is the length of the longest path from an input bit to the output.
\end{definition}

The depth of a circuit is an important characteristic, especially in the context of parallel computation, where it corresponds to the time neccessary to evaluate the circuit (with a sufficient number of processes).

\vspace{3mm}
\begin{center}
\includegraphics[width=50mm]{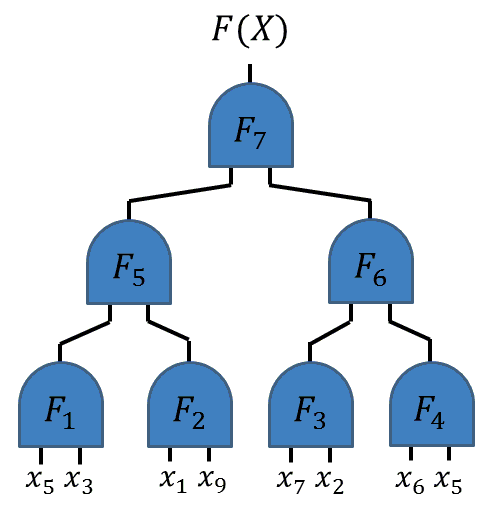}\\
A depth 3 clasifier $F$ made of 7 2-gates for data $x=(x_1,x_2,\ldots,x_{10})$.
\end{center}
\vspace{3mm}

To describe this circuit we need to specify the the inputs for each gate (which may be some dimensions of the input, or outputs of other gates), as well as filling in 7 truth tables (one for each gate).

Suppose now that we wish to evaluate this classifier on 64000 different examples (10 bits each).
Then the dataset can be stored as a table with 64000 rows of 10 bits each. 

The first step ef every algorithm will always be to transpose the input. In this example we obtain a table with 10 lines of 64000 bits.
On a 64 bit system this requires 1000 memory words per lines.

Now assume that all the gates $F_i$ are either AND, OR, or XOR gates. 
Evaluating this classifier on that data will require evaluating the 7 binary gates. 
Since the gates we have chosen can be evaluated by any reasonable system on words of 64 bits, 
the evaluation will require 7 vector operations on 1000 words tables, for a total of 7000 word operations. 
Note that here, 7000 is significantly smaller the 640,000 bits (the total size of the data). 

In general we will want to use more then 7 gates and more importantly might want to work with gates of arity larger then 2. 
Fortunately the AND, OR and NOT gates are more then enough to compute any computable function. 
We exploit this fact to develop a general technique for evaluating circuits with gates of any arity.

In this framework, we consider the input bits as (unlearned) features, and the output of the gates, $F_i$ as learned features.  
For each feature, we store its negation as well, to save operations.
This representation will be useful in the implementation of gates. 
The input bit will be the first feature and can be compute by on vector operation (NEGATION).
The tensor product of a list of $k$ bits is the list containing $2^k$ being the product (logical AND) of all possible combination of every input and its negation.

\begin{center}
  \begin{tabular}{|| c|c|| c|c|| c|c||c|c|c|c|c|c|c|c|| }
    \hline\hline
       $x_0$&$x_1$&$y_0$&$y_1$&$z_0$&$z_1$&$v_0$&$v_1$&$v_2$&$v_3$&$v_4$&$v_5$&$v_6$&$v_7$\\ \hline
       0&1&  0&1& 0&1&  0&  0&0&  0&0&  0&0&  1 \\ \hline
       0&1&  0&1& 1&0&  0&  0&0&  0&0&  0&0&  1 \\ \hline
       0&1&  1&0&  0&1& 0&  0&0&  0&0&  1&0&  0 \\ \hline
       1&0&  0&1&  0&1& 0&  0&0&  1&0&  0&0&  0 \\ \hline
       1&0&  1&0&  1&0& 1&  0&0&  0&0&  0&0&  0 \\ \hline
    \hline
  \end{tabular}
   \captionof{table}{$x \otimes y \otimes z=v$ for five training examples}
\end{center}

If the truth table of a gate gives us its output for each input then the tensor product of the feature given as input to the gates have a one to one corespondent with the truth table.
The output of a gate is a feature where the first bit is the OR of every element of the tensor product position where the truth table ha a value 0 and the second vector of the output of the gate is the OR of all tensor product element where the truth table has a 1. 
Using the fact that the output feature is composed of a vector and its complement one can spare some operations.

\begin{lemma}
On an architecture with words of $m$ bits, the evaluation of a $k$-gate when the size of the data set is $n$ (a multiple of $m$) is $(2^k+(2^{(k-1)}+1)) n / m$ elementary gates. 
\end{lemma}

On one core, using no parallelism (other then the fact that 32-bit words are used), using C\# we can evaluate 27 million 4-gates over 32 bits. This means that a 5000 gate circuit with 5000 inputs can be evaluated in a second. Of course parallelism can be used both at the vector level and the circuit level, for significant speedups.

%%%%%%%%%%%%%%%%%%%%%%%%%%%%%%%%%%%%%%%%%%%%%%%%%%%%%%%%%%%%%%%%%%%%%%%%%%%%%%%%%%%%%%%%%%%%%%%%%%%%%%%%%%%%%%%%%%%%%%
%%%%%%%%%%%%%%%%%%%%%%%%%%%%%%%%%%%%%%%%%%%%%%%%%%%%%%%%%%%%%%%%%%%%%%%%%%%%%%%%%%%%%%%%%%%%%%%%%%%%%%%%%%%%%%%%%%%%%%
\section*{Greedy algorithm}

The best classifier would be the {\em simplest} circuit capable of calssifying the dataset. 
This optimization is obviously intractable.  
We simplify this optimization problem several times, obtaining a greedy algorithm that produces a good and extremely efficient classifier.  
First, we set the arity of all gates to be equal to $k$ (first hyperparameter of our algorithm). 
Second, we set the topology of our circuit to be a full tree of depth $d$ (the second hyperparameter of our algorithm).
Third, we let the leaves' inputs each be a random coordinate of the input space.
Finaly, we use greedy local optimization instead of global optimization.

The greedy algorithm is quite simple to understand. 
Every leaf takes as input a random coordinate (a bit) of the binary input vector. 
To specify the circuit we then have to specify the truth table of each of its gates.
 
We chose these truth tables in a greedy way starting with the gates connected to the leaf (data bits) and climbing up the tree until the root gate, which outputs the classification decision. 
So how do we chose the truth tables for the non-root gates? 
In fact, every gate simply behaves as if its output were output of the entire circuit, i.e., the classification decision.
Every individual gate will be chosen to be the {\em best} gate to accomplish this task, i.e. to minimize the classification error of its outputs. 
The reader might recall that the number of different gates of arity $k$ is double exponential in $k$ and might worry that this optimization is intractable. 

That optimization can be simplfied, as follows:
In the case of a $k$-gate, we have $2^k$ possible inputs to worry about and we have to specify the answer of the gate on each of those, nothing less, but also nothing more. 

We can optimize a gate independently for each possible input vector, e.g. the best answer for a 4-gate on input $(0,0,0,0)$ can be computed independently to the best outcome on input $(0,0,0,1)$ and so on. 

This means for a $k$-gate we only need to compute $2^k$ values, not $2^{2^k}$. 
This is reasonable to do when $k$ is small (we mostly use gates of size 2 to 8 in this work). 

We can compute the best output for a given input simply by counting how many examples that produce this input for this gate belong to each category.  
If there are more such examples belonging to category 0, we chose that element of the truth table to be 0, otherwise, we set it to 1.

We start this process of specifying the gates one by one with the gates at the bottom of the tree, which take dimensions of training data as input.
Once these gates' truth tables are known, we also know their output on each example, which we use to compute the second row's gates' truth tables, and so on, moving up the tree until we have learned all the gates.

\subsection*{Maximizing information gain versus Maximizing probability}

Early on in our investigation, we decided that maximizing the success probability is only a good criterion for the top gate.
For lower gates, one should maximize the information gain (Kullback–Leibler divergence).

\begin{definition}
Information gain: $D(P||Q)=\sum_i \left(\frac{P(i)}{Q(i)}\right) P(i)$
\end{definition}

This can also be done efficiently.  
For any given gate, we can calculate the input to this gate on any data example.  
Now, for each possible input, $v$, to the gate, $F_i$, we compute the proportion of examples producing $v$ as input to $F_i$ also are ultimately classified as category 0.
We can then sort the $2^k$ possible inputs to $F_i$ based on this proportion, creating a list $l$.  
Then we find the two set partition respecting that order which maximizes the information gain (i.e., we choose an index, $j$, of $l$ and set the output of $F_i$ to be 0 for all previous indices, and 1 for $j$ and all subsequent indices).
This leads to the optimal gate.

\subsection*{Experimental results}

We perform experiments with a variety of data, ranging from black and white images, and RGB images of common objects, to numeric data. 
In every experiment, the data is encoded as binary vectors, with the order of dimensions playing no role (note that this makes the algorithm permutation-invariant). 

The two hyperparameters of the algorithm are the arity of gates ($k$) and the depth of the circuit ($d$).
The running time scales exponentially with both hyperparameters, but in practice very good result can be obtain with small values of both $k$ and $d$. 
Arity of 4 and depth of 8 appears to be a good default hyperparameter setting that gave good results in all our experiments. 
With this configuration, the classifier is a tree with 21845 internal nodes and 65536 leaves. For those hyperparameter the C\# training time (32 bits) is roughly 10.6s when the number of training example is 50,000 (Using 6 cores and 12 treads the time is reduced to 2.5s). 
Note that the time required to evaluate the classifier is not much smaller; more than a second is needed to performs 50000 classifications, with such a circuit.  
With default hyperparameters our algorithm compares favorably with neural nets and with the right choice of hyperparameters it is possible to obtain even better results. 

%%%%%%%%%%%%%%%%%%%%%%%%%%%%%%%%%%%%%%%%%%%%%%%%%%%%%%%%%%%%%%%%%%%%%%%%%%%%%%%%%%%%%%%%%%%%%%%%%%%%%%%%%%%%%%%%%%%%%%%%%%%%%%%%%%%%%%%%%%%%%%%%%%%%%%%%%%%%%%%%%%%%%%%%%%%%%%%%%%%%%%%%%%%%%%%%%%%%%%%%%%%%%%%%%%%%%%%%%%%%%%%%%%%%%%%%%%%%
\section*{Main algorithm}

We have observed that the previous greedy algorithm result is highly dependent on choice of leaves' inputs.
Repeating independently at random with different configurations of inputs is not very efficient. 
In this section we propose to use simple hill climbing to optimize the choices of input dimension for all leaves. 
This technique is quite powerful and if not used carefully, can result in severe over-fitting.

Again, the idea of this technique is quite simple. 
Chose one leaf uniformly at random and change it to take a different input dimension, also chosen uniformly at random. 
If the new classifier is better keep it like that and if not, go back the how it was before.
Of course, this process must be repeated an appropriate number of times.

\begin{figure}
\begin{center}
\includegraphics[width=50mm]{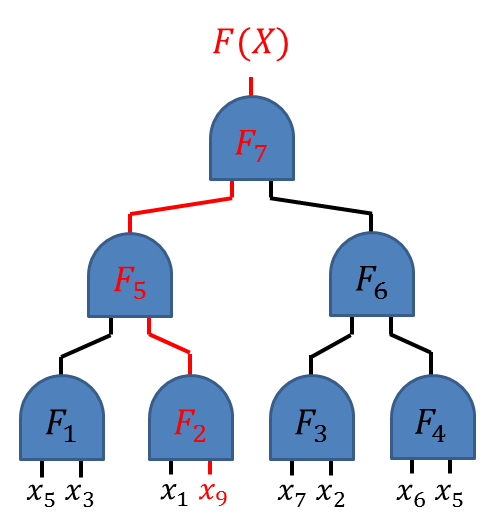}
\end{center}
\caption{Update scheme of a depth 3 classifier $F$}
\label{hill_climbing}
\end{figure}

This is efficient, because only the gates along the path from that leaf to the root need to be re-evaluated.
For a $k$ array tree this value is the base-$k$ logarithm of the number of leaves and even for very large trees this value is quite small.
For example, with the default hyperparameters, only 7 nodes need to be recomputed. (See figure \ref{hill_climbing})

\subsection*{Experimental results}

This algorithm allows us to obtain classifiers that are significantly smaller then the greedy algorithm produces, resulting in very significant improvement for a given tree size. 
For example, we are able to classify the data using only a few gates. 
On at least one data set, we obtain results that are significantly better than the 2007 state of the art as presented in \cite{Larochelle:2007}

%%%%%%%%%%%%%%%%%%%%%%%%%%%%%%%%%%%%%%%%%%%%%%%%%%%%%%%%%%%%%%%%%%%%%%%%%%%%%%%%%%%%%%%%%%%%%%%%%%%%%%%%%%%%%%%%%%%%%%
%%%%%%%%%%%%%%%%%%%%%%%%%%%%%%%%%%%%%%%%%%%%%%%%%%%%%%%%%%%%%%%%%%%%%%%%%%%%%%%%%%%%%%%%%%%%%%%%%%%%%%%%%%%%%%%%%%%%%%
\section*{Conclusion and future work}

In this report we have presented a new paradigm for machine learning.
Our preliminary results that strongly indicate that the paradigm is worth studying.
To do so we have presented a simple but dramatically new machine learning algorithm that is very efficient and learns good binary classifiers. 
Experimental results seems to indicate that the technique performs significantly better then feed-forward neural networks with a single hidden layer and that it can outperform deep neural nets in some contexts, as well. 

Here is a partial list of things to investigate for this method in the context of supervised classification.
Although the preliminary results we have obtained are impressive, we believe further investigation is needed, and that the technique can be improved significantly.

\begin{itemize}
\item Our experiments were performed on 6 CPU corse, but using GPUs should enable even faster implementation.
\item FPGAs should also make computation faster.  Indeed, they appear taylor-made for this algorithm, with an ultra-fast physical implementation of circuits composed of 6-gates. 
      We are investigating their use on classification problems, and our result will be presented in a coming report. 
\item Taking into consideration the structure of the data for the initialization of the leaves in the circuit tree could lead to significant improvement, especially when $n=0$.
\item Preprocessing of the data will certainly yield improvement. This can be done in several ways with or without exploiting the structure of the input. Dimensions with very low entropy could be eliminated and dimensions that are very similar replace by a single dimenstion that is their average.  
\item Adding noise in part of the learning process could reduce over-fitting.
\item We can perform hill climbing on the nodes themselves once a good tree is obtained. This looks very promising.
\item We can compute more features than needed and select a subset with low mutual information.
\item Explore with more example the advantage/disadvantages of using binary representations for the data.
\item In every algorithm, the top of the tree tends to be composed of gates that simply output to mode of their inputs. 
      We might be able to improve on this by using something else for the top levels.
\item Can we extend this framework to perform efficient, effective multiclass classification? 
\end{itemize}

We believe that deep learning of representations using autoencoders can be accomplish in this framework.
That being said, the techniques will certainly share some properties with those based on neural nets (hopefully the impressive results), where we already know that there is a significant difference between discrete and continuous cases.
For example, using binary outputs, there is no gradient and things like hamming distance are a poor substitute.
In some future work we will investigate an autoencoder and we are very optimistic this will leads to interesting results.

%%%%%%%%%%%%%%%%%%%%%%%%%%%%%%%%%%%%%%%%%%%%%%%%%%%%%%%%%%%%%%%%%%%%%%%%%%%%%%%%%%%%%%%%%%%%%%%%%%%%%%%%%%%%%%%%%%%%%%
\subsection*{Acknowledgment}

Thanks to David Krueger for useful comments. 
A special thanks to François Leclerc who significantly contributed to this work by its expertise in programing and with helpful comments. 
I would also like to take this opportunity to thanks Yoshua Bengio and his group for accepting me as an intern form my 6 month sabbatical. 
Most of this work was completed prior to the sabbatical. 
Finally, I would also like to thanks my wife Sandra for her unconditional support.

\bibliographystyle{alpha}	  % (uses file "plain.bst")
\bibliography{ML}		% expects file "myrefs.bib"
\nocite{*}

%%%%%%%%%%%%%%%%%%%%%%%%%%%%%%%%%%%%%%%%%%%%%%%%%%%%%%%%%%%%%%%%%%%%%%%%%%%%%%%%%%%%%%%%%%%%%%%%%%%%%%%%%%%%%%%%%%%%%%
%%%%%%%%%%%%%%%%%%%%%%%%%%%%%%%%%%%%%%%%%%%%%%%%%%%%%%%%%%%%%%%%%%%%%%%%%%%%%%%%%%%%%%%%%%%%%%%%%%%%%%%%%%%%%%%%%%%%%%
%%%%%%%%%%%%%%%%%%%%%%%%%%%%%%%%%%%%%%%%%%%%%%%%%%%%%%%%%%%%%%%%%%%%%%%%%%%%%%%%%%%%%%%%%%%%%%%%%%%%%%%%%%%%%%%%%%%%%%
\newpage\appendix

\section*{Appendix: Detailed experimental results}

The algorithm we present in this note is a general purpose supervised learning algorithm for binary classification. 
As stated, the data to be classified are binary vectors (i.e. strings of bits) and we have avoided preprocessing or using domain-knowledge in order to investigate 
the intrinsic power of the technique as a black-box, permutation-invariant algorithm.  Although we experiment with images, our method completely ignores the spatial structure.

Here is a list of the variables (with typical values) used in the following experimental tests:
\begin{itemize}
\item Training = 12k, 12,000 examples for training
\item Test = 50k, 50,000 examples for testing
\item $a=4$, arity of the gates, 4-gates
\item $d = 8$, dept of the tree
\item $t$ is the dept up to which update are propagated in the hill climbing
\item $n=0$, number of trials for hill climbing
\item $\delta = 0.1$, noise level (only for the CUBE dataset)
\item Bits = 2, number of bits of precision per pixel; means 2 bit by color channel for color pictures and 2 bit in total for black and white pictures
\end{itemize}

In our experiments, we noticed that the greedy algorithm with arity 4 and depth 8 always gave relatively good performance. 
We refer to these as the default hyperparameters, or defaults.
The running time of the algorithm using the defaults does not depend on the data type but only on the number of training examples. 
For 10k examples the algorithm completes its training in less then a second. 
If $n=0$ we recover the greedy algorithm, otherwise this describes the hill climbing algorithm.

It seems one of the main advantage of the classification algorithm we have presented here is it's efficiency. 
All the experiment we done on a 6 core Intel Core i7 CPU with a sufficient (64GB) amount of memory. 
The algorithm is trivially parallelized, and we ran 12 threads in parallel for most experiments.
The presented timing results are with a C\# version of the algorithm; Python code will be made available.
The fact that both python and C\# enable binary vector operation very efficiently makes the algorithm run extremely fast.

We have asked a student to performs some tests using FPGA. 
The structure of the FPQA uses 6-gates with a fully programmable truth table.
The FPGA will not be used for training but it is perfectly suited for classification.
Detailed experimental results will be presented in another report.  Our preliminary results indicate that 500k classifications per second are possible even with a large classifier (several tens of thousands of leaves). 
No experiments have been performed with GPU accelerated computing, but we are optimistic that this would yield the expected speedup. 

%%%%%%%%%%%%%%%%%%%%%%%%%%%%%%%%%%%%%%%%%%%%%%%%%%%%%%%%%%%%%%%%%%%%%%%%%%%%%%%%%%%%%%%%%%%%%%%%%%%%%%%%%%%%%%%%%%%%%%
\subsection*{CUBES}

\begin{figure}
\begin{center}
\includegraphics[width=130mm]{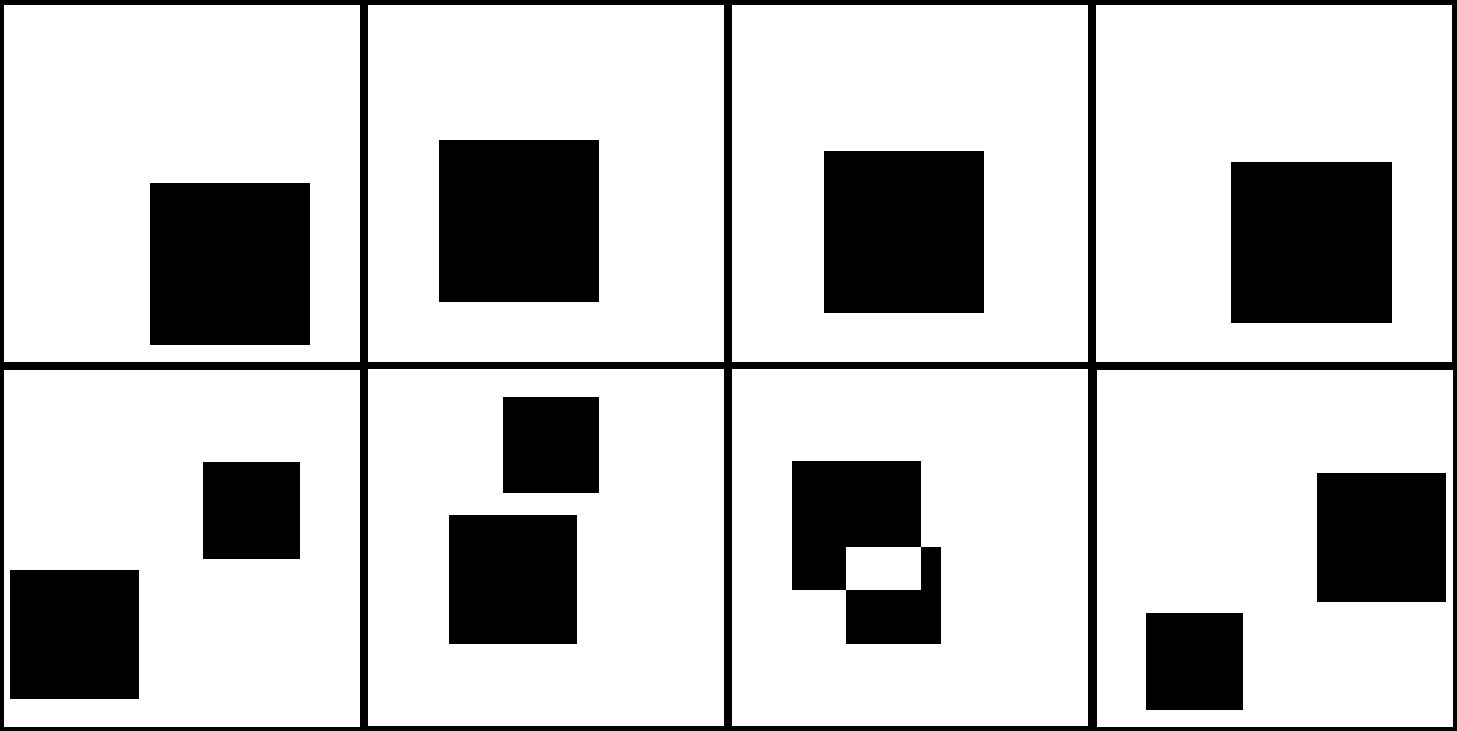}\\
\bf		8 examles at $\delta=0$\% \\
\vspace{6mm}
\includegraphics[width=130mm]{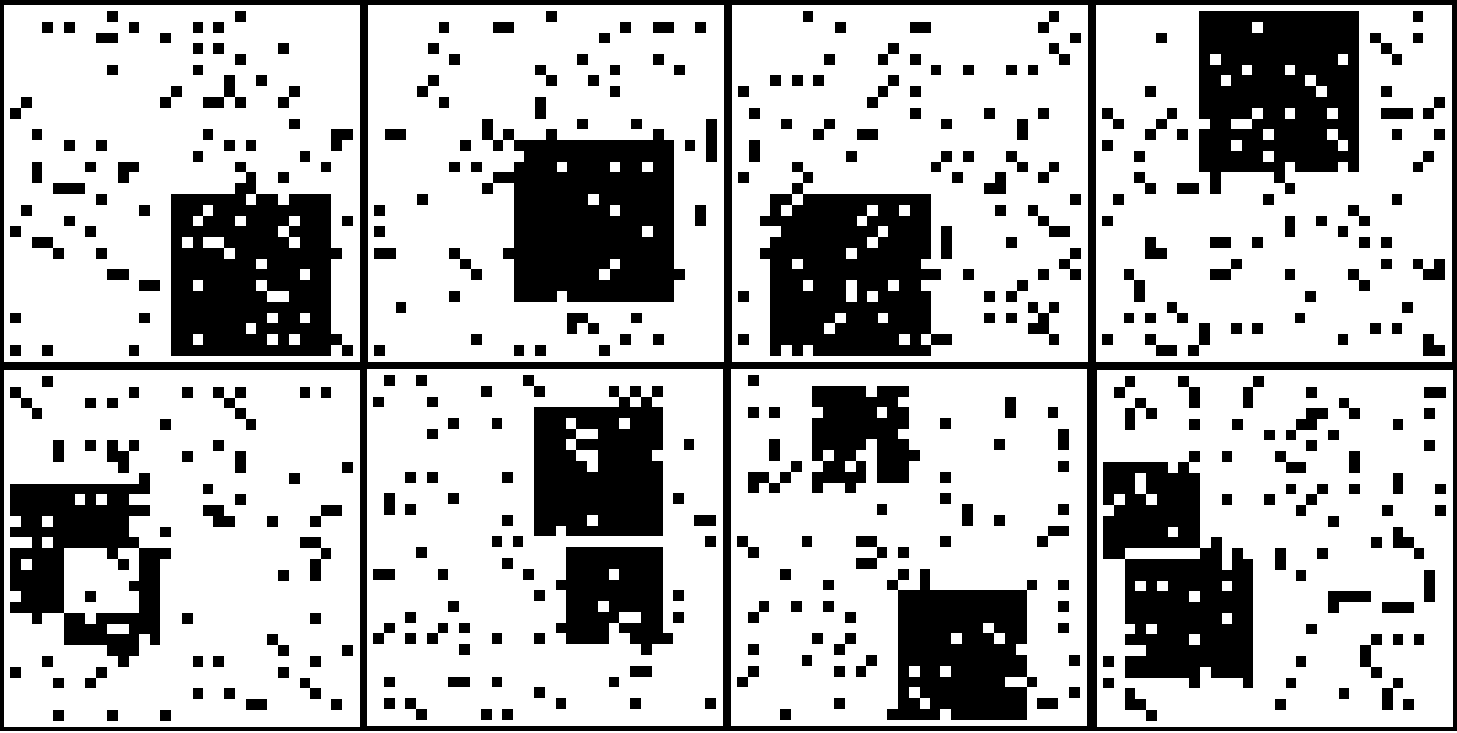}\\
\bf		8 examles at $\delta=10$\% \\
\vspace{6mm}
\includegraphics[width=130mm]{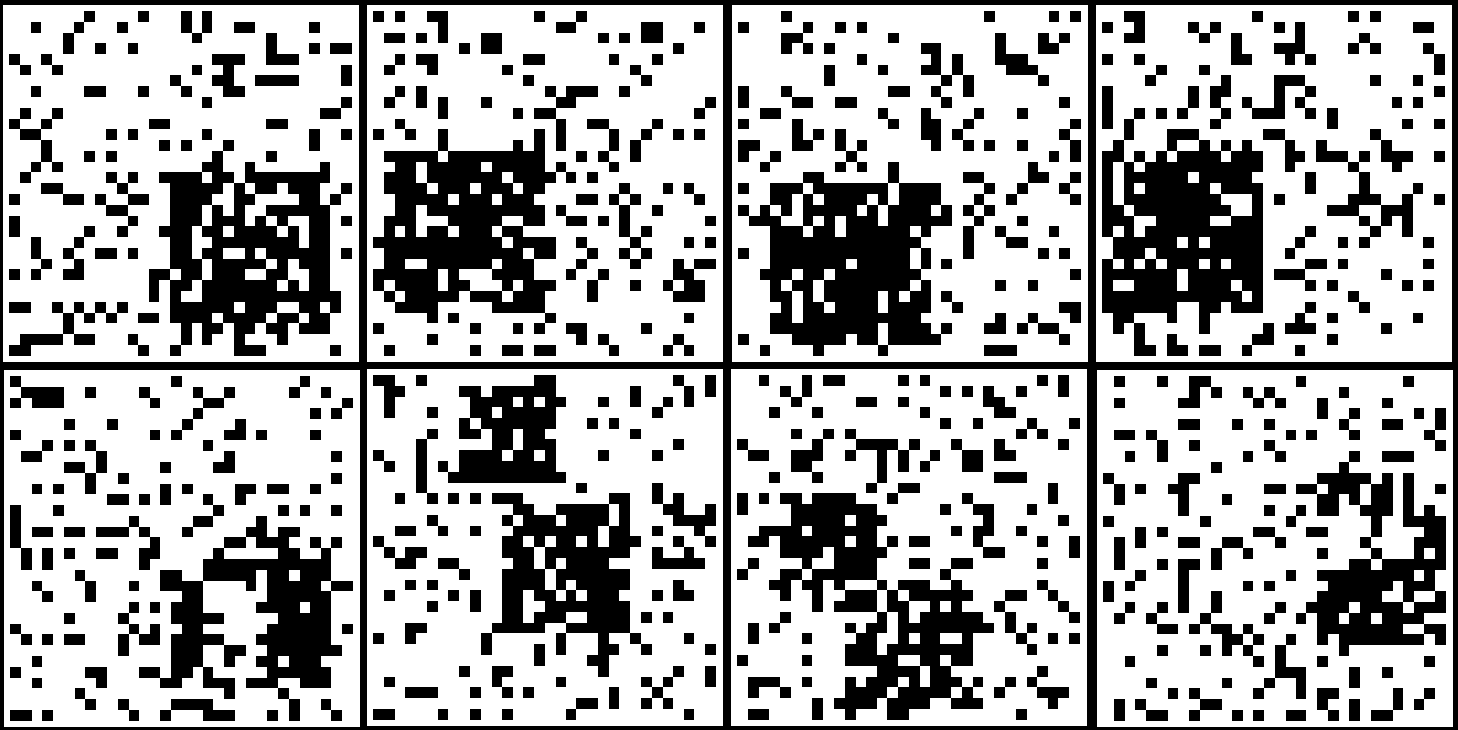}\\
\bf		8 examles at $\delta=20$\% \\
\end{center}
\caption{CUBE}
\label{figure_cube}
\end{figure}

The first dataset we use is artificially constructed, so a practically infinite number of examples can be constructed.
The CUBE data set (figure \ref{figure_cube}) is composed of images of size 32 by 32 of which the pixel are black or white 
(1024 bits string). 
The first class contain a black cube of 15 by 15 at a random location and the second class contains two cubes, one 12 by 12 and one 9 by 9. 
When those two cubes do not overlap they have in both case a surface of 225 bits as a single 15 by 15 cube. 
When the cubes overlap we chose the overlap to be white. 
In this and all cases, the permutation-invariant nature of the algorithm means it does not have acces to information regarding the relative locations of pixels.
This data set have a single parameter which is the level of added noise (between 0 and 1). 
For a noise $\delta$ every bit is independently flipped with probability $\delta$.

\begin{figure}
\begin{center}
  \begin{tabular}{| l |l|| c|c|c|c|c|c|c|c|c|c|c|| }
    \hline
    $\delta=0.0$ &        &      &     &       &    &    &    &    &    &   & &    \\ \hline\hline
    $a \backslash d$ &         &  2     &  4    & 6   & 8     & 10   & 12   & 14    & 16  & 18   &  20  & 22  \\ \hline\hline
		2     & Train   & 43.5   & 34.2 & 24.9 & 25.5  & 16.0 & 14.1 & 10.9  & 9.0 & 7.6  &  6.4 & 5.0  \\ \hline
    2     & Test    & 43.6   & 34.1 & 24.6 & 25.2  & 15.3 & 14.2 & 11.1  & 9.1 & 7.8  &  6.2 & 5.2  \\ \hline\hline
		3     & Train   &  37.1  & 19.2 & 13.6 & 8.08  & 3.93 & 2.08 & 1.78  &     &      &      &      \\ \hline
    3     & Test    &  37.0  & 19.0 & 14.0 & 8.15  & 3.97 & 2.05 & 2.06  &     &      &      &    \\ \hline\hline
		4     & Train   &  27.8  & 10.6 & 2.37 & 0.62  & 0.12 &      &       &     &      &      &    \\ \hline
    4     & Test    &  27.7  & 10.8 & 2.60 & {\bf 0.72}  & 0.21 &      &       &     &      &      &    \\ \hline\hline
		$a \backslash d$ &         &  2     & 3    & 4    & 5     & 6    & 7    &       &     &      &      &    \\ \hline\hline
    5     & Train   &  32.0  & 18.3 & 4.82 & 0.57  & 0.00 & 0.00 &       &     &      &      &    \\ \hline
    5     & Test    &  32.4  & 18.4 & 4.92 & 0.51  & 0.02 & 0.17 &       &     &      &      &    \\ \hline\hline 
		6     & Train   &  18.9  & 5.28 & 0.19 & 0.00  & 0.00 &      &       &     &      &      &    \\ \hline
    6     & Test    &  18.8  & 5.31 & 0.23 & {\bf 0.03}  & 0.20 &      &       &     &      &      &    \\ \hline\hline
		7     & Train   &  14.4  & 0.26 & 0.00 &       &      &      &       &     &      &      &    \\ \hline
    7     & Test    &  14.1  & 0.24 & 0.13 &       &      &      &       &     &      &      &    \\ \hline\hline
		8     & Train   &  11.6  & 0.18 & 0.00 &       &      &      &       &     &      &      &    \\ \hline
    8     & Test    &  11.7  & 0.46 & 0.20 &       &      &      &       &     &      &      &    \\ \hline\hline
		9     & Train   &  3.52  & 0.00 &      &       &      &      &       &     &      &      &    \\ \hline
    9     & Test    &  4.15  & 0.55 &      &       &      &      &       &     &      &      &    \\ \hline\hline
    10    & Train   &  3.49  & 0.00 &      &       &      &      &       &     &      &      &    \\ \hline
    10    & Test    &  5.04  & 1.18 &      &       &      &      &       &     &      &      &    \\ \hline\hline
		11    & Train   &  3.38  & 0.00 &      &       &      &      &       &     &      &      &    \\ \hline
    11    & Test    &  6.79  & 3.33 &      &       &      &      &       &     &      &      &    \\ \hline \hline\hline
		$\delta=0.1$ &        &      &     &       &    &    &    &    &    &   & &    \\ \hline\hline
    $a \backslash d$ &        &  4    & 6    & 8    & 10   & 12   & 14   & 16   & 18   & 20         &      &      \\ \hline\hline
    2     & Train  &  38.4 & 34.8 & 33.2 & 28.6 & 25.1 & 24.1 & 23.4 & 22.9 & 23.3      &      &       \\ \hline
    2     & Test   &  38.6 & 35.3 & 33.8 & 29.3 & 25.8 & 24.7 & 24.1 & 23.6 & 24.1      &      &       \\ \hline\hline
		4     & Train  &  26.1 & 18.6 & 15.4 & 15.7 &      &      &      &      & & &           \\ \hline
    4     & Test   &  27.1 & 19.4 & 17.0 & 17.3 &      &      &      &      & & &           \\ \hline\hline
		$a \backslash d$ &        &  2    & 3    & 4    & 5    & 6    & 7    &      &      & & &           \\ \hline\hline
		6     & Train  &  30.8 & 21.5 & 15.0 & 10.4 & 9.42 & 8.53 &      &      & & &           \\ \hline 
    6     & Test   &  32.8 & 23.4 & 17.3 & 12.9 & 12.7 & 12.1 &      &      & & &           \\ \hline \hline
		8     & Train  &  23.2 & 10.7 & 5.32 & 3.23 & 2.14 & 1.57 &      &      & & &           \\ \hline
    8     & Test   &  26.7 & 15.1 & 10.9 & 8.94 & {\bf 8.52} & 8.78 &      &      & & &           \\ \hline\hline
    10    & Train  &  14.1 & 2.99 & 0.12 & 0.00 &      &      &      &      &      &      &             \\ \hline
    10    & Test   &  23.3 & 14.2 & 11.7 & 17.0 &      &      &      &      &      &      &             \\ \hline\hline
  \end{tabular}
   \captionof{table}{CUBES, $\delta= 0.0,0.1$, $a =$ 2-10, $d = $ 2-20, $n =$ 0, Training  = 12k, Test = 50k}
\end{center}
\end{figure}

\begin{center}
  \begin{tabular}{|l|l||c|c|c|c|c|c|c|c|c|c|c|| }
    \hline
		$a=2$   &        &     &     &     &     &   & & & &       \\ \hline\hline
    $d \backslash t$    &        & 1    & 2    & 3    & 4    & 5    & 6    & 7    & 8    &  9       \\ \hline\hline
    4     & Train  & 23.6 & 18.8 & 11.6 & 4.38 &      &      &      &      &          \\ \hline
    4     & Test   & 23.8 & 18.9 & 11.9 & 4.55 &      &      &      &      &          \\ \hline\hline
		6     & Train  &      & 10.3 & 6.62 & 2.23 & 0.83 &      &      &      &          \\ \hline
    6     & Test   &      & 10.5 & 6.63 & 2.38 & 1.00 &      &      &      &          \\ \hline\hline
		8     & Train  &      & 10.0 & 5.32 & 2.72 & 0.06 & 0.16 & 0.00 &      &          \\ \hline
    8     & Test   &      & 10.6 & 5.14 & 3.06 & 0.10 & 0.27 & 0.01 &      &          \\ \hline\hline
    10    & Train  &      & 9.50 & 5.08 & 2.92 & 1.11 & 0.00 & 0.00 & 0.00 &  0.00    \\ \hline
    10    & Test   &      & 9.57 & 5.11 & 2.88 & 1.41 & 0.00 & 0.02 & 0.35 &  0.25    \\ \hline\hline
		$a=4$   &        &     &     &     &     &         & & & &   \\ \hline\hline
		$d \backslash t$ &        & 1    & 2    & 3    & 4    & 5      & & & &     \\ \hline\hline
    2     & Train  & 8.63 & 1.06 &      &      &       & & & &         \\ \hline
    2     & Test   & 9.07 & 1.12 &      &      &      & & & &          \\ \hline\hline
		4     & Train  & 3.33 & 0.00 & 0.00 & 0.00 &       & & & &        \\ \hline
    4     & Test   & 3.58 & {\bf 0.00} & 0.41 & 0.41 &         & & & &        \\ \hline\hline
		6     & Train  &      & 0.00 & 0.00 & 0.00 & 0.00    & & & &     \\ \hline
    6     & Test   &      & 0.02 & 0.15 & 1.62 & 0.36   & & & &    \\ \hline\hline
  \end{tabular}
   \captionof{table}{CUBES, $\delta= 0.0$, $a = $2-4, n=100k, Training = 12k, Test = 50k}
\end{center}

\begin{center}
  \begin{tabular}{|l|l||c|c|c|c|c|c|c|c|c|c|c|| }
    \hline
     $d \backslash t$     &        & 1    & 2    & 3    & 4    & 5    & 6    & 7           \\ \hline\hline
    4     & Train  & 17.2 & 11.4 & 8.76 & 7.23 &      &      &             \\ \hline
    4     & Test   & 17.9 & 12.5 & 11.1 & 12.7 &      &      &             \\ \hline\hline
		6     & Train  &      & 10.3 & 5.56 & 3.94 & 4.23 &      &             \\ \hline
    6     & Test   &      & 11.5 & 7.37 & {\bf 6.25} & 8.73 &      &             \\ \hline\hline
		8     & Train  &      & 13.8 & 10.5 & 8.45 & 7.64 & 8.01 & 8.16        \\ \hline
    8     & Test   &      & 15.2 & 12.4 & 10.6 & 10.6 & 12.3 & 12.9        \\ \hline\hline
  \end{tabular}
   \captionof{table}{CUBES,$\delta = 0.1$, a = 4, n=100k, Training = 12k, Test = 50k}
\end{center}

When the noise is 20\% ($\delta=0.2$) the default hyperparameter algorithm gives errors of $24.6$ for training and 27.4 on the test set.

\begin{center}
  \begin{tabular}{|l|l||c|c|c|c|c|c|c|c|c|c|c|| }
    \hline
    $n$    &       & 0    & 50k  & 100k & 500k & 1m   & 2m          \\ \hline\hline
    4     & Train  & 16.8 & 13.1 & 11.8 & 8.85 & 7.85 & 7.17        \\ \hline
    4     & Test   & 23.3 & 21.2 & 19.8 & 17.4 & 16.5 & {\bf 15.9}  \\ \hline\hline
  \end{tabular}
   \captionof{table}{CUBES, $\delta= 0.2$, $a = 6$, $d=6$, $t=3$, $n=$ 0-2m, Training = 12k, Test = 50k}
\end{center}

In term of running time, the C\# version on the algorithm with default hyperparameters trains in less then a second ($a = 4$, $d=8$, $t=4$, $n=$0, Training = 12k).
The running time of the algorithm with defaults depends only on the number of examples, it does not even depends on size of each example.
Its running time on datasets of a million example is roughly 30 seconds.
The longest training example of this section took less then 6 minutes 
($a = 6$, $d=6$, $t=3$, $n=$2m, Training = 12k).
The python version is slower by a factor 3, approximately. 

%%%%%%%%%%%%%%%%%%%%%%%%%%%%%%%%%%%%%%%%%%%%%%%%%%%%%%%%%%%%%%%%%%%%%%%%%%%%%%%%%%%%%%%%%%%%%%%%%%%%%%%%%%%%%%%%%%%%%%
\subsection*{GAUSS: Numbers with a normal distribution}

The second dataset is also computer generated. 
That being said it is very different from the previous one.
In this dataset the examples are lists of 32 integers (16 bits/integer) generated from one of two normal distributions.
The task is to classify example lists according to which distribution their integers are drawn from.  
Since the algorithm does not have access to the structure of the input, it receives no information about which bits are part of which integer. 
We performed experiments in which the two distributions differ in mean (top two plots) and variance (bottom two plots).

\begin{figure}
\begin{center}
\includegraphics[width=110mm]{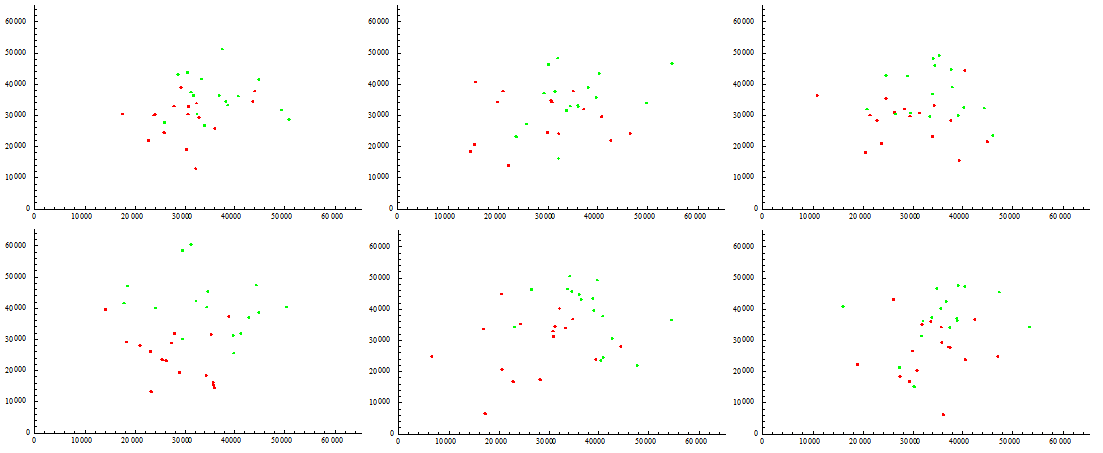}\\
\bf	$(\mu=28768, \sigma=8000)$ versus $(\mu=28768+8000, \sigma=8000)$
\vspace{4mm}
\includegraphics[width=110mm]{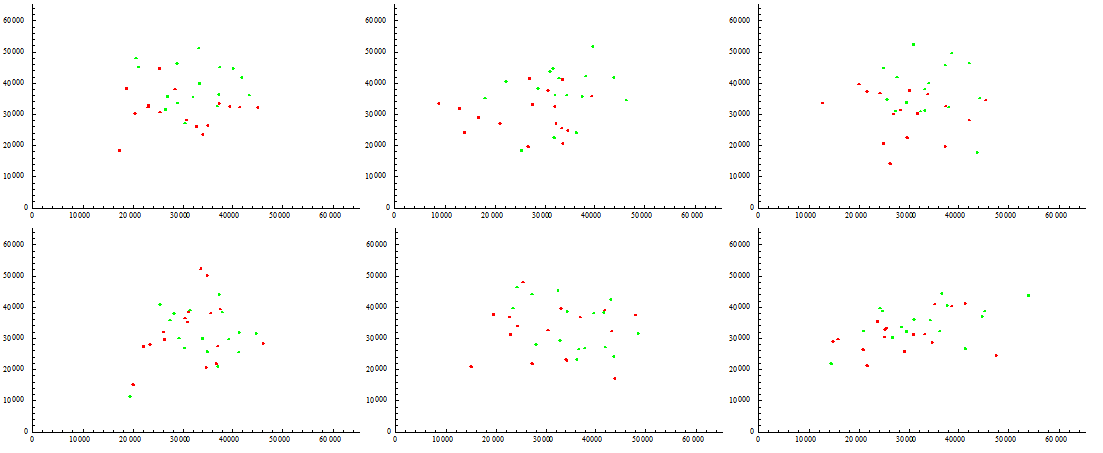}\\
\bf	$(\mu=30768, \sigma=8000)$ versus $(\mu=30768+4000, \sigma=8000)$
\vspace{4mm}
\includegraphics[width=110mm]{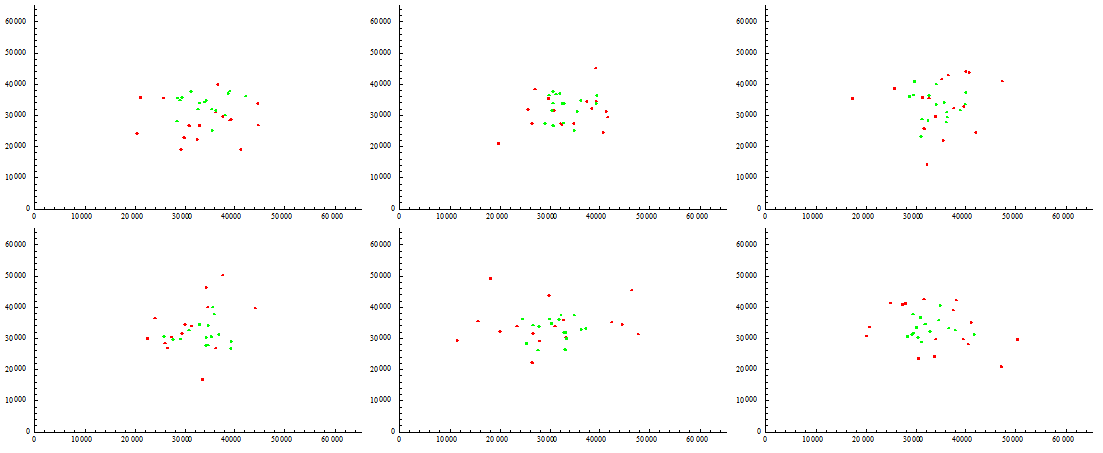}\\
\bf	$(\mu=32768, \sigma=2000)$ versus $(\mu=32768, \sigma=8000)$
\vspace{4mm}
\includegraphics[width=110mm]{Normal_var_2.png}\\
\bf	$(\mu=32768, \sigma=4000)$ versus $(\mu=32768, \sigma=8000)$
\end{center}
\captionof{figure}{}
\label{GAUSSdata}
\end{figure}

We also performed experiments to see if the scale has an impact on the algorithm; the results we obtained show that apart from rounding errors and significant overflows, rescaling all data by a constant factor does not significantly affect the performance of the classifiers. 
Figure \ref{GAUSSdata} shows a graphical representation of the different datasets we used. 
For visualization, we have interpreted the data as 16 pairs of number which we plot in table \ref{GAUSS_result1} and \ref{GAUSS_result2}.

\begin{center}
  \begin{tabular}{|l|l||c|c|c|c|c|c|c|c|c|c|c|| }
    \hline
          &       & 0    & 100k  & 500k            \\ \hline\hline
     $(\mu=28768, \sigma=8000)$                  & Train  & 1.67 & 1.30 & 0.96      \\ \hline
     versus $(\mu=28768+8000, \sigma=8000)$     & Test   & 2.11 & 1.70 & 1.61      \\ \hline\hline
	   $(\mu=30768, \sigma=8000)$                  & Train  & 14.4 & 13.0 & 11.6      \\ \hline
     versus $(\mu=30768+4000, \sigma=8000)$     & Test   & 15.4 & 15.0 & 14.4      \\ \hline\hline
		 $(\mu=32768, \sigma=2000)$                  & Train  & 0.14 & 0.00 & 0.00      \\ \hline
      versus $(\mu=32768, \sigma=8000)$         & Test   & 0.19 & 0.09 & 0.06      \\ \hline\hline				
		 $(\mu=32768, \sigma=4000)$                  & Train  & 5.90 & 2.05 & 1.31      \\ \hline
     versus $(\mu=32768, \sigma=8000)$     			& Test   & 15.2 & 5.55 & 1.97      \\ \hline\hline
  \end{tabular}
   \captionof{table}{GAUS,  $a = 4$, $d=8$, $t=4$, $n=$ 0-500k, Training = 10k, Test = 10k}
\label{GAUSS_result1}
\end{center}

\begin{center}
  \begin{tabular}{|l|l||c|c|c|c|c|c|c|c|c|c|c|| }
    \hline
          &       & 0    & 100k  & 500k            \\ \hline\hline
     $(\mu=28768, \sigma=8000)$                  & Train  & 1.07 & 0.72 &       \\ \hline
     versus $(\mu=28768+8000, \sigma=8000)$     & Test   & 2.25 & 1.86 &       \\ \hline\hline
	   $(\mu=30768, \sigma=8000)$                  & Train  & 9.85 & 8.92 &  7.99 \\ \hline
     versus $(\mu=30768+4000, \sigma=8000)$     & Test   & 15.5 & 15.2 &  14.7¸\\ \hline\hline
		 $(\mu=32768, \sigma=2000)$                  & Train  & 0.02 & 0.00 &       \\ \hline
      versus $(\mu=32768, \sigma=8000)$         & Test   & 0.54 & 0.06 &      \\ \hline\hline				
		 $(\mu=32768, \sigma=4000)$                  & Train  & 1.05 & 0.20 &       \\ \hline
     versus $(\mu=32768, \sigma=8000)$     			& Test   & 28.2 & 5.10 &       \\ \hline\hline
  \end{tabular}
   \captionof{table}{GAUSS,  $a = 6$, $d=6$, $t=3$, $n=$ 0-500k, Training = 10k, Test = 10k}
\label{GAUSS_result2}
\end{center}

%%%%%%%%%%%%%%%%%%%%%%%%%%%%%%%%%%%%%%%%%%%%%%%%%%%%%%%%%%%%%%%%%%%%%%%%%%%%%%%%%%%%%%%%%%%%%%%%%%%%%%%%%%%%%%%%%%%%%%
\subsection*{MNIST: digits 3 versus 5}

Using the famous MNIST dataset we perform binary classification on the set of 3s and 5s.  
After several experiments, we found these classes the hardest to distinguish. 
For example, distinguishing 0 and 1 with very high probability only require {\rm five} 4-gates and driving the test error to zero is very straightforward.
For this dataset the only hyperparameter is the number of bits for the pixel representation. 
Each pixel is an 8 bit integer ranging form 0 to 255. With $k$ bits of resolution we chose the first $k$ most significant bits.
We have obtained very good results with this dataset. 
The fact that the algorithm gives good result on 8 bit resolution does not imply that it behaves well on binarized MNIST (1-bit resolution), but rather that the algorithm do not loose its power in presence of a large number of non significant bits. 
With default hyperparameters, the greedy algorithm achieved a test error of 5.6\%, and using hill climbing is it possible to drive down the error even more, to 3.57\%. 
The results are presented in table \ref{MNIST_result1},\ref{MNIST_result2} and \ref{MNIST_result3}.

\vspace{4mm}
\begin{center}
\includegraphics[width=120mm]{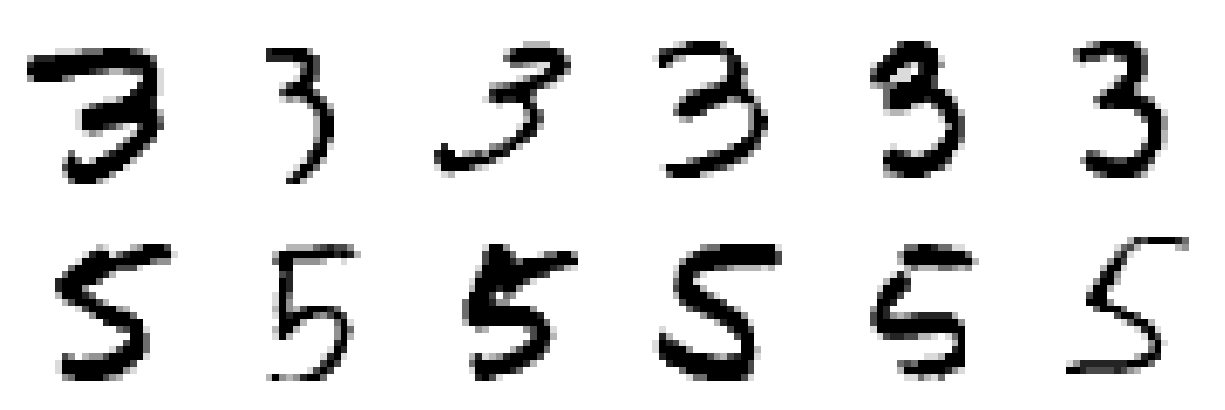}\\
\bf	MNIST 8 bits per pixel
\end{center}

\vspace{4mm}

\begin{center}
  \begin{tabular}{| l |l|| c|c|c|c|c|c|c|c|c|c|c|| }
    \hline		
    1 bit &        &      &     &       &    &    &    &    &    &       \\ \hline\hline
    $a \backslash d$ &        &  4    & 6    &  8     & 10   & 12   & 14   & 16   & 18   &  20     \\ \hline\hline
    2     & Train  &  30.7 & 21.0 &  13.2  & 10.3 & 8.22 & 6.02 & 6.02 & 5.62 &  5.14   \\ \hline
    2     & Test   &  31.2 & 23.8 &  14.5  & 12.5 & 9.63 & 7.37 & 6.82 & 7.08 &  6.93   \\ \hline\hline
		4     & Train  &  11.8 & 5.67 &  3.21  & 2.72 & 1.93 &      &      &      &      \\ \hline
    4     & Test   &  14.2 & 7.21 &  {\bf 5.57}  & 5.37 & 5.37 &      &      &      &      \\ \hline\hline\hline
    $a \backslash d$ &        &  2    & 3    &  4    & 5     & 6    & 7    &      &      &        \\ \hline\hline\hline
		6     & Train  &  25.2 & 7.47 &  4.66 & 2.15  & 0.88 & 0.48 &      &      &         \\ \hline 
    6     & Test   &  28.4 & 9.37 &  8.04 & 5.87  & 5.84 & 6.21 &      &      &         \\ \hline \hline
		8     & Train  &  20.4 & 3.47 &  0.66 & 0.04  & 0.00 &      &      &      &     \\ \hline
    8     & Test   &  24.7 & 9.48 &  9.07 & 14.5  & 21.6 &      &      &      &     \\ \hline\hline
    2 bits &        &      &     &       &    &    &    &    &    &       \\ \hline\hline
    $a \backslash d$ &         &  4    & 6    &  8     & 10   & 12   & 14    & 16   & 18   &  20     \\ \hline\hline
    2     & Train   &  26.7 & 24.2 &  17.8  & 14.8 & 8.44 & 7.29  & 6.68 & 5.36 &  5.14   \\ \hline
    2     & Test    &  28.0 & 27.0 &  20.1  & 15.8 & 9.49 & 7.48  & 7.95 & 6.94 &  6.39   \\ \hline\hline
		4     & Train   &  11.8 & 4.79 &  3.73  & 2.72 & 1.89 &       &      &      &      \\ \hline
    4     & Test    &  13.4 & 6.52 &  {\bf 6.23}  & 5.61 & 5.62 &       &      &      &      \\ \hline\hline\hline
		$a \backslash d$ &         &  2    & 3    &  4    & 5     & 6    & 7     &      &      &   \\ \hline\hline\hline
		6     & Train   &  22.1 & 12.4 & 4.39  & 2.50  & 1.41 & 0.48  &      &      &   \\ \hline 
    6     & Test    &  26.7 & 16.6 & 8.03  & 5.99  & 6.00 & 6.42  &      &      &   \\ \hline \hline
		8     & Train   &  21.0 & 3.87 & 0.70  & 0.04  & 0.00 &       &      &      &  \\ \hline
    8     & Test    &  26.1 & 10.8 & 9.72  & 15.6  & 24.2 &       &      &      &  \\ \hline\hline
    8 bits &        &      &     &       &    &    &    &    &    &       \\ \hline\hline
    $a \backslash d$ &         &  4    & 6    &  8     & 10   & 12   & 14    & 16   & 18   &  20     \\ \hline\hline
    2     & Train  & 41.4 & 22.1 & 18.8 & 14.6 & 9.58 & 8.35 & 6.90 & 6.77 &  5.62   \\ \hline
    2     & Test   & 39.8 & 24.1 & 18.7 & 15.8 & 10.9 & 10.3 & 8.21 & 8.24 &  7.16  \\ \hline\hline
		4     & Train  & 13.4 & 6.46 & 4.13 & 3.16 &      &      &      &      &      \\ \hline
    4     & Test   & 14.9 & 8.91 & {\bf 6.87} & 6.01 &      &      &      &      &      \\ \hline\hline\hline
    $a \backslash d$ &        & 2    & 3    &  4   & 5    & 6    & 7    &      &      &           \\ \hline\hline\hline
		6     & Train  & 28.9 & 14.8 & 5.58 & 2.99 & 1.98 & 0.66 &      &      &          \\ \hline 
    6     & Test   & 31.9 & 18.0 & 9.00 & 7.11 & 6.42 & 7.87 &      &      &          \\ \hline \hline
		8     & Train  & 14.9 & 7.21 & 1.10 & 0.04 & 0.00 &      &      &      &      \\ \hline
    8     & Test   & 19.8 & 15.5 & 11.2 & 14.2 & 21.9 &      &      &      &      \\ \hline\hline
  \end{tabular}
   \captionof{table}{MNIST,Bits = 1-8,$a= $2-9, $d=$2-20,  Training = 2276, Test = 9662}
	\label{MNIST_result1}
\end{center}

\begin{center}
  \begin{tabular}{|l|l||c|c|c|c|c|c|c|c|c|c|c|| }
    \hline
          &        & 2    & 4    & 6    & 8    & 10    & 12    & 14    & 8    &  9       \\ \hline\hline
    4     & Train  & 11.8 & 4.79 &      &      &      &      &      &      &          \\ \hline
    4     & Test   & 13.0 & 7.96 &      &      &      &      &      &      &          \\ \hline\hline
		8     & Train  & 8.04 & 3.51 & 1.93 &      &      &  &  &      &          \\ \hline
    8     & Test   & 9.17 & 4.81 & 5.31 &      &      &  &  &      &          \\ \hline\hline
		12    & Train  & 6.37 & 4.83 & 2.90 & 1.41 & 0.83 &  &  &  &      \\ \hline
    12    & Test   & 8.46 & 6.58 & 4.71 & 4.43 & 5.22 &  &  &  &     \\ \hline\hline
		16    & Train  & 6.50 & 5.67 & 4.22 & 2.90 & 2.72 & 1.76 & 1.93 &  &      \\ \hline
    16    & Test   & 7.64 & 6.84 & 6.02 & 5.29 & 5.60 & 5.07 & 7.43 &  &     \\ \hline\hline	
\end{tabular}
   \captionof{table}{MNIST,$a = 2$, Bits = 1, $n=$100k, Training = 2276, Test = 9662}
\label{MNIST_result2}
\end{center}

\begin{center}
  \begin{tabular}{|l|l||c|c|c|c|c|c|c|c|c|c|c|| }
    \hline
          &        & 1    & 2    & 3    & 4    & 5    & 6    & 7    & 8    &  9       \\ \hline\hline
		2     & Train  & 9.40 & 4.48 &      &      &      &      &      &      &          \\ \hline
    2     & Test   & 11.1 & 7.72 &      &      &      &      &      &      &          \\ \hline\hline
    4     & Train  & 5.01 & 1.98 & 0.57 & 0.18 &      &      &      &      &          \\ \hline
    4     & Test   & 5.89 & 3.91 & 4.04 & 5.36 &      &      &      &      &          \\ \hline\hline
		6     & Train  &      & 2.28 & 1.05 & 0.26 & 0.09 &      &      &      &          \\ \hline
    6     & Test   &      & 4.95 & {\bf 3.57} & 4.70 & 4.47 &      &      &      &          \\ \hline\hline
		8     & Train  &      & 2.68 & 1.80 & 1.27 & 0.88 & 0.97 & 0.75 &      &          \\ \hline
    8     & Test   &      & 5.30 & 4.71 & 4.21 & 4.64 & 5.10 & 5.85 &      &          \\ \hline\hline
  \end{tabular}
   \captionof{table}{MNIST,Bits = 1,$a = 4$,  $n=$100k, Training = 2276, Test = 9662}
\label{MNIST_result3}
\end{center}

%%%%%%%%%%%%%%%%%%%%%%%%%%%%%%%%%%%%%%%%%%%%%%%%%%%%%%%%%%%%%%%%%%%%%%%%%%%%%%%%%%%%%%%%%%%%%%%%%%%%%%%%%%%%%%%%%%%%%%
\subsection*{CONVEX}

This dataset was created by the LISA group in Montreal and it is a hard classification problem for deep feedforward nets.
The data consists of binary images, whose 0-valued pixels form either a convex or non-convex set, as shown in figure \ref{convex_images}.  The classification problem is to determine if the set is convex or not.  
On this dataset, we obtained significantly better results than the 2007 state of the art, as presented in \cite{Larochelle:2007}.
The size of the training dataset used by \cite{Larochelle:2007} was 8k and the test dataset has 50k elements. 
We used the same number of examples for our experiments. 

The algorithms compared on this dataset are: Support vector machine with RBF kernel (SVM RBF), support vector machine with polynomial kernel (SVM Poly), feed-forward neural network (NNET) with a single hidden layer, deep believe network with 3 hidden layers (DBN-3), stacked autoassociator with 3 hidden layers (SAA-3). 
We have included in the table the performance of our algorithm with 3 settings of the hyperparameters.
{\em Tapp A} uses the default hyperparameters, {\em Tapp B} sets $a=8$, $d=7$, and $n=0$ (hill climbing is not used), and finally the best performing algorithm {\em Tapp C}, had $a = 6$ $d= 7$, $t = 3$ and  $n = 20m$.  
Training error of 4.94\% was achieved, but with significant over-fitting, as the test error was $15.9$.
We did not perform a thorough hyperparameter search, and we expect, e.g., an increase in running time to allow further improvement. 
{\em Tapp C} took much longer to train than any other result we present in this article (2 hours, compared to {\em Tapp A} running is less then a second).

\begin{center}
  \begin{tabular}{|c|c|c|c|c|c|c|c|c|c|c|c|| }
    \hline
	SVM  RBF	& SVM  Poly	& NNet	& DBN-3	& SAA-3	&  Tapp A &  Tapp B & Tapp C\\ \hline
  19.13 &	19.82  &	32.25  &	18.63  & 	18.41  &  29.9 & 21.5 & {\bf 15.89 }  \\ \hline
  \end{tabular}
   \captionof{table}{CONVEX, Training = 8k, Test = 50k}
	 \label{convex_images}
\end{center}

\vspace{4mm}
\begin{center}
\includegraphics[width=120mm]{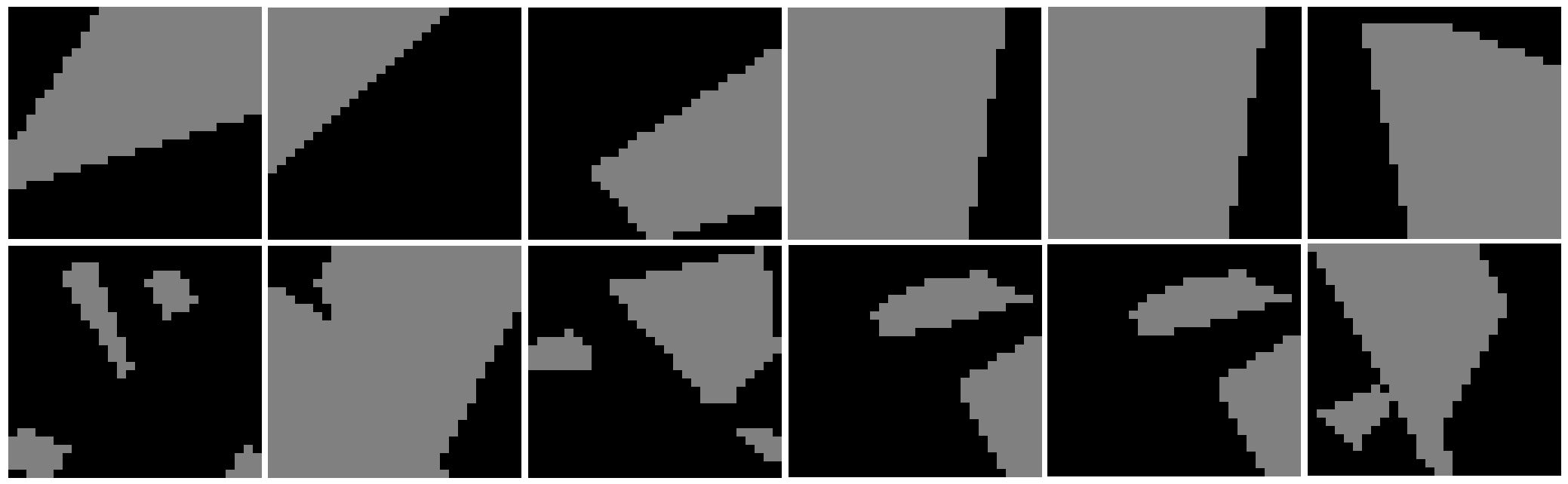}\\
{\bf		CONVEX \\  First line is convex and second line is non convex}
\end{center}
\vspace{4mm}

\begin{center}
  \begin{tabular}{|l|l||c|c|c|c|c|c|c|c|c|c|c|| }
    \hline
    $a \backslash d$ &        & 4    & 6    & 8    & 10   & 12   & 14   & 16   & 18   &  20      \\ \hline\hline
    2     & Train  & 47.5 & 45.4 & 41.9 & 40.7 & 37.1 & 37.3 & 37.2 & 35.2 &  35.3    \\ \hline
    2     & Test   & 47.8 & 46.2 & 43.1 & 42.0 & 39.2 & 39.1 & 39.3 & 38.3 &  38.4    \\ \hline\hline
		3     & Train  & 41.6 & 37.5 & 32.8 & 31.6 & 30.3 & 29.4 &      &      &          \\ \hline
    3     & Test   & 43.4 & 38.5 & 34.8 & 34.9 & 34.0 & 33.7 &      &      &          \\ \hline\hline
		4     & Train  & 34.9 & 29.3 & 26.7 & 25.5 &      &      &      &      &          \\ \hline
    4     & Test   & 36.9 & 32.3 & {\bf 29.9} & 30.7 &      &      &      &      &          \\ \hline\hline\hline
    $a \backslash d$ &        & 2    & 3    & 4    & 5    & 6    & 7    & 8    &      &          \\ \hline\hline\hline
    5     & Train  & 41.5 & 35.6 & 30.3 & 26.3 & 22.9 & 21.4 & 20.8 &      &          \\ \hline
    5     & Test   & 42.8 & 37.2 & 32.8 & 29.0 & 26.5 & 26.0 & 26.2 &      &          \\ \hline\hline
		6     & Train  & 39.8 & 30.7 & 23.1 & 20.5 & 17.9 & 16.6 & 15.9 &      &          \\ \hline 
    6     & Test   & 41.1 & 33.0 & 26.9 & 24.9 & 23.2 & 23.6 & 23.1 &      &          \\ \hline \hline
		7     & Train  & 37.4 & 25.0 & 19.8 & 15.9 & 14.2 & 12.6 & 11.1 &      &          \\ \hline
    7     & Test   & 38.7 & 29.0 & 24.9 & 23.2 & 22.2 & 22.2 & 21.9 &      &          \\ \hline\hline
		8     & Train  & 33.2 & 22.3 & 14.9 & 11.9 & 9.60 & 7.76 &      &      &          \\ \hline
    8     & Test   & 34.8 & 27.9 & 23.2 & 22.2 & 21.7 & {\bf 21.5} &      &      &          \\ \hline\hline
		9     & Train  & 31.0 & 16.5 & 11.3 & 7.95 & 5.55 &      &      &      &          \\ \hline
    9     & Test   & 33.8 & 24.8 & 22.7 & 21.7 & 22.6 &      &      &      &          \\ \hline\hline
		10    & Train  & 24.7 & 11.9 & 7.01 & 4.09 &      &      &      &      &          \\ \hline
    10    & Test   & 29.9 & 24.0 & 22.6 & 24.6 &      &      &      &      &          \\ \hline\hline
  \end{tabular}
   \captionof{table}{CONVEX, $a =$ 2-9, Training = 8k, Test = 50k}
\end{center}

\vspace{7mm}
%%%%%%%%%%%%%%%%%%%%%%%%%%%%%%%%%%%%%%%%%%%%%%%%%%%%%%%%%%%%%%%%%%%%%%%%%%%%%%%%%%%%%%%%%%%%%%%%%%%%%%%%%%%%%%%%%%%%%%
\subsection*{CIFAR 10}

To show the performance of our algorithm on RGB natural images we chose two categories of the CIFAR 10 image dataset: cars and birds.

\vspace{4mm}
\begin{center}
\includegraphics[width=120mm]{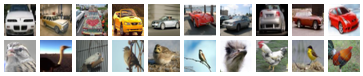}\\
\bf	CIFAR 10, cars versus birds
\end{center}
\vspace{4mm}

The original images have 8 bit per color channel with a total of 24 bits per pixel but using more then 4 bit per channel does not provide much visual improvement.
All experiments used 10k training and 2k testing images. 
With default hyperparameters, the algorithm had a test error of 19.2.  With hill climbing, the test error can easily be reduces to less then 15\%. 

\begin{center}
  \begin{tabular}{| l |l|| c|c|c|c|c|c|c|c|c|c|c|| }
    \hline
   $a \backslash d$ &         &  8    & 9    &  10       \\ \hline\hline
    1     & Train   & 18.4  & 18.1 &  17.7     \\ \hline
    1     & Test    & 20.1  & 21.2 &  20.6     \\ \hline\hline
    2     & Train   & 16.7  & 15.7 &  16.0     \\ \hline
    2     & Test    & 18.6  & 18.6 &  15.4     \\ \hline\hline
    3     &  Train  & 16.4  & 16.3 &  16.2     \\ \hline
    3     &  Test   & 19.7  & 18.5 &  18.8     \\ \hline\hline
    4     &  Train  & 17.5  & 16.7 &  16.2     \\ \hline
    4     &  Test   & 20.5  & 19.2 &  18.5     \\ \hline\hline
    8     & Train   & 18.0  & 16.9 &  16.7     \\ \hline
    8     & Test    & 19.6  & 19.2 &  19.1     \\ \hline\hline
    \end{tabular}
   \captionof{table}{CIFAR, a=4, $d=$ 8-10, Bits = 1-8, Training  = 10k, Test = 2k}
\end{center}

\begin{center}
  \begin{tabular}{| l ||l| c|c|c|c|c|c|c|c|c|c|c|| }
    \hline
          $n=$     & 100k  & 200k &  400k  & 800k          \\ \hline\hline
          Train   & 14.1  & 13.3 &  12.4  & 11.6        \\ \hline
          Test    & 17.3  & 16.6 &  15.9  & 15.6      \\ \hline
    \end{tabular}
   \captionof{table}{CIFAR, $a = $4, $d=9$, $t=3$, Bits = 2, $n=$100k to 800k, Training = 10k, Test = 2k}
\end{center}

\begin{center}
  \begin{tabular}{| l ||l| c|c|c|c|c|c|c|c|c|c|c|| }
    \hline
           $n=$      & 0     & 21.6k & 43.2k & 86.4k & 172k & 432k           \\ \hline\hline
           Train   & 12.2  & 10.9  & 10.2  & 9.66  & 8.97 & 8.35     \\ \hline
           Test    & 17.0  & 16.1  & 15.4  & 14.5  & 14.4 & 14.3     \\ \hline
    \end{tabular}
   \captionof{table}{CIFAR, $a = 6$, $d=6$, $t=3$, Bits $= 2$, $n=$100k to 800k, Training = 10k, Test = 2k}
\end{center}

\end{document}